%% file: main.tex
\documentclass[runningheads]{llncs}

\usepackage[font=footnotesize,labelfont=bf]{caption}
\usepackage{subcaption}
\usepackage{tikz}
\usepackage{comment}
\usepackage{amsmath,amssymb}
\usepackage{color}
\usepackage{wrapfig}
\usepackage{multirow}
\usepackage[accsupp]{axessibility}

\usepackage{authblk}
\usepackage{marvosym}

\input{00-format-and-definitions}

\begin{document}
\pagestyle{headings}
\mainmatter
  
\def\ECCVSubNumber{1234}  

\title{HiCAST: Highly Customized Arbitrary Style Transfer with Adapter Enhanced Diffusion Models} 

\newcommand{\algoName}{HiCAST\xspace}


\author{
Hanzhang Wang \inst{1} \footnote[4]{Work done when Hanzhang Wang is a Research Intern in Baidu Research} \footnote[5]{Equal contribution} \and
Haoran Wang\inst{2} $^\ddagger$ \footnote[6]{Corresponding authors} \and
Jinze Yang\inst{4} \and
Zhongrui Yu\inst{5} \and
Zeke Xie\inst{2} \and
Lei Tian\inst{3} \and
Xinyan Xiao\inst{3} \and
Junjun Jiang\inst{1} \and
Xianming Liu\inst{1} $^\S$ \and 
Mingming Sun\inst{2}
}

\institute{
$^1\,$Harbin Institute of Technology,
$^2\,$Cognitive Computing Lab, Baidu Research,
$^3\,$Baidu Inc.,
$^4\,$University of Chinese Academy of Sciences,
$^5\,$ETH Zurich \\
\email{\{wanghaoran09, xiezeke, tianlei09, xiaoxinyan, sunmingming01\}@baidu.com,\\
\{cswhz, jiangjunjun, csxm\}@hit.edu.cn, yangjinze20@mails.ucas.ac.cn, zhonyu@ethz.ch},
}

\maketitle


\input{0-abstract}
\input{1-introduction}
\input{2-related-work}
\input{3-method}
\input{4-experiments}

\input{5-conclusion}

\small
\bibliographystyle{splncs04}
\bibliography{main.bbl}

\newpage

\input{6-appendix}

\end{document}

%% file: 00-format-and-definitions.tex
\usepackage{graphicx}
\usepackage{epsfig}

\usepackage{subcaption}
\usepackage{float}
\usepackage[font=small]{caption}
\usepackage{lscape}      
\usepackage{etoolbox} 

\usepackage[ruled, vlined, linesnumbered]{algorithm2e}

\usepackage{microtype}
\usepackage[utf8]{inputenc} 
\usepackage[T1]{fontenc}    
\usepackage{microtype}
\DisableLigatures{encoding = *, family = *}

\usepackage{bm}                          
\usepackage{mathtools}
\usepackage{amsmath, amssymb, amscd}
\usepackage{wasysym} 
\usepackage{nicefrac}       
\usepackage{mdwmath}
\usepackage{eqparbox} 
\usepackage{array} 
\usepackage{nicefrac}

\usepackage{array} 
\usepackage{tabularx}
\usepackage{multirow}
\usepackage{multicol}
\usepackage{booktabs}   

\usepackage{listings}
\usepackage{nameref}
\usepackage{paralist}
\usepackage{enumitem}
\usepackage{xspace}
\usepackage{setspace}

\usepackage[numbers, sort]{natbib}

\usepackage{graphicx}
\usepackage{subcaption}
\usepackage[font=small]{caption}

\usepackage{float}
\usepackage{lscape}

\usepackage{booktabs}
\usepackage{multirow}

\usepackage{paralist}
\usepackage{enumitem}

\usepackage{bm}
\usepackage{epsfig}
\usepackage{mathtools}

\usepackage{color,colortbl}
\usepackage{siunitx} 

\usepackage{comment}

\usepackage{url}  
\usepackage[pagebackref=true,breaklinks=true,colorlinks,urlcolor=blue,citecolor=blue,linkcolor=blue,bookmarks=false]{hyperref}

\usepackage{listings}

\usepackage{xspace}
\usepackage{setspace}

\usepackage{nicefrac}
\usepackage{microtype}
\usepackage[utf8]{inputenc} 
\usepackage[T1]{fontenc}    

\usepackage{url}  
\usepackage{color,colortbl} 

\definecolor{Lightapricot}{rgb}{0.99,0.84,0.69}

%% file: 0-abstract.tex
\begin{abstract}

The goal of Arbitrary Style Transfer (AST) is injecting the artistic features of a style reference into a given image/video. Existing methods usually focus on pursuing the balance between style and content, whereas ignoring the significant demand for flexible and customized stylization results and thereby limiting their practical application. To address this critical issue, a novel AST approach namely HiCAST is proposed, which is capable of explicitly customizing the stylization results according to various source of semantic clues. In the specific, our model is constructed based on Latent Diffusion Model (LDM) and elaborately designed to absorb content and style instance as conditions of LDM. It is characterized by introducing of \textit{Style Adapter}, which allows user to flexibly manipulate the output results by aligning multi-level style information and intrinsic knowledge in LDM. Lastly, we further extend our model to perform video AST. A novel learning objective is leveraged for video diffusion model training, which significantly improve cross-frame temporal consistency in the premise of maintaining stylization strength. Qualitative and quantitative comparisons as well as comprehensive user studies demonstrate that our HiCAST outperforms the existing SoTA methods in generating visually plausible stylization results.

\end{abstract}

%% file: 1-introduction.tex
\section{Introduction}

In recent years, style transfer \cite{gatys2016image,an2021artflow,liu2021adaattn,deng2022stytr2} has become a popular research topic in computer vision, which refers to integrating the content of one image/video with the unique stylistic patterns of a given artistic work. Concretely, it can be categorized into three types \cite{jing2019neural}: Per-Style-Per-Model methods, Multiple-Style-Per-Model methods and Arbitrary-Style-Per-Model approaches. The last one, also called Arbitrary Style Transfer (AST), aims at transferring arbitrary artistic styles once upon one single model is trained. In contrast to two others, it is a more advantageous and versatile solution, which has become more appealing for both academic and industrial communities. 

As a seminal work, Gatys et al. \cite{gatys2016image} propose an optimization-based approach. It used a pre-trained deep neural network to extract features from both content and style, followed by iteratively minimizing the joint content and style loss in the feature space. Although it is very time-consuming, this work motivates a rich line of subsequent studies \cite{huang2017arbitrary,li2019learning,an2021artflow,liu2021adaattn,wu2021styleformer,deng2022stytr2}. In early stage, numerous researchers focus on utilizing convolutional neural networks (CNNs) to learn style and content representations. However, the convolution operations in CNNs lead to limited receptive field, thereby are not adept at capturing long-range dependencies. Moreover, some details will be lost when the depth of CNNs continuously increases, and consequently degrade the results \cite{deng2022stytr2}. To ameliorate this issue, more advanced representation modules, such as attention mechanism \cite{liu2021adaattn} or transformer \cite{wu2021styleformer,deng2022stytr2}, are introduced to improve stylization results. 

Despite recent progress, there still exists a non negligible gap between the existing synthesized stylizations and ideal AST product. In brief, most prevalent AST methods are built based on an \textit{encoder-transfer-decoder} framework: an encoder extracting content and style features, followed by a transformation module to adjust features statistics, and finally a decoder generating stylization results. Specifically, the encoders are usually pre-trained on visual recognition tasks, such as VGG, and the other two modules are trained from scratch for AST. It is notable none of them are pre-trained on visual generation tasks, consequently limiting the ability to get more vivid and creative results. Furthermore, owing to only being trained on content and style data pair, the decoders lack enough generalization ability for handling some Out-of-Data (OOD) cases. 

Recently, Diffusion Models (DMs) \cite{ho2020denoising} emerged and rapidly receive a surge of interest. With the emergence of large-scale image-text database \cite{srinivasan2021wit,schuhmann2022laion}, DMs have displayed their mighty visual generation ability, which also benefit a wide range of related applications \cite{saharia2022image,lugmayr2022repaint,brempong2022denoising,zhou20213d,chen2022analog}. Naturally, the success of DMs attracts more researchers to investigate in \textit{how to exploit DMs to improve AST results}. The corresponding works can be classified into two categories: 1) training-free methods \cite{kwon2022diffusion} and 2) training-based methods \cite{ruta2023diff,chen2023artfusion}. The former is achieved by designing specific objective to guide the reverse diffusion process without model fine-tuning, which is efficient to implement but mainly dependent on the capacity of employed DMs. In comparison, the training-based methods are orthogonal to the former and have greater potential. Specifically, Ruta at el. \cite{ruta2023diff} proposed to integrate style guidance into the self-attention blocks of the trained LDM Unet modules, which only fine-tunes the attention blocks. Concurrently, Chen at el. \cite{chen2023artfusion} propose to inject content and style image into the LDM model as conditions, in which the whole model is trained from scratch. It is able to control the stylization strength of results by adjusting the weights of classifier-free guidance. Nevertheless, we raise a question here: \textit{Is it ideal enough for a AST product that manipulates the stylization results on only one general level namely strength} ? 

It is well known that, the quality of human-created artworks is evaluated according to human perception, which include multiple dimensions: such as brushstrokes, color distributions, textures, semantics, etc. Only adjusting the stylization strength on a general level can not satisfy our need for AST. 
To be specific, we further decompose the holistic information of creation into two levels: low level (e.g., textures, edges) and high level (e.g., structure, semantics). It is imperative to endow AST model with the abiltiy to customize the stylization results according to explicit clues from various semantic levels. Although recent literatures have made impressive progress, this critical problem is still unexplored and remained. 

To tackle this problem, we draw inspiration from the emergence of the DMs guided by additional conditions, such as ControlNet \cite{zhang2023adding} and T2I-Adapter \cite{mou2023t2i}. The success of them motivates us to consider:  \textit{Can we also plug the external signal into the DMs through model training, so as to customize the stylization results in explicit manner ?} 

\begin{figure}[!t]
    \centering
    \includegraphics[width=0.9\linewidth,height=4.6cm]{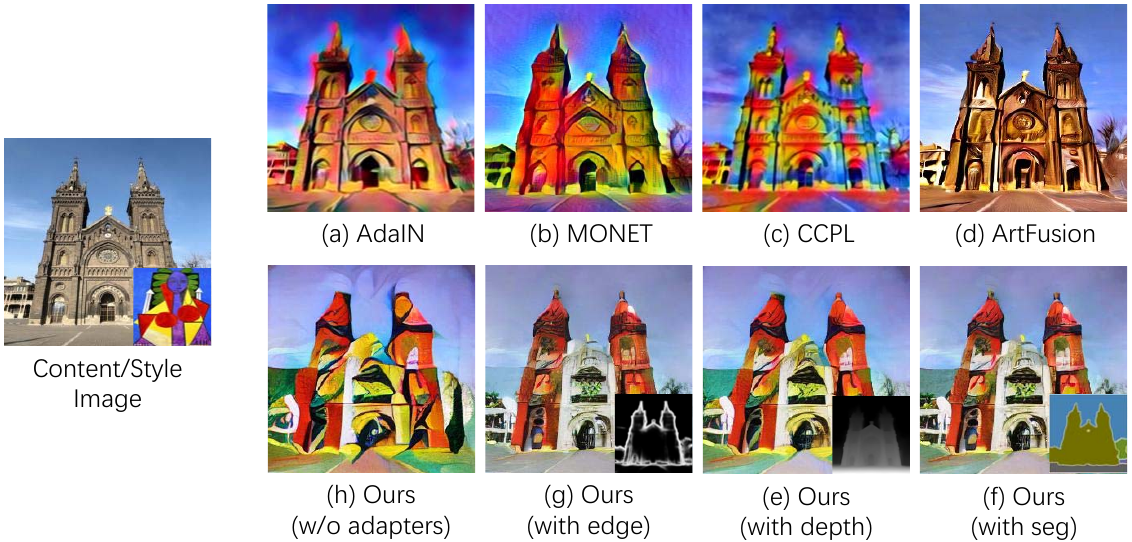}
    \caption{In contrast to existing AST methods (a-d), our proposed HiCAST model can customize the stylization results according to different control signals (e.g. edge (f), depth (g), semantic segmentation (h)). [Best viewed with zooming-in]}
    \label{fig:figure1}
    \vspace{-20pt}
\end{figure}

To achieve this goal, we propose a novel AST approach namely \textbf{Hi}ghly \textbf{C}ontrollable \textbf{A}rbitary \textbf{S}tyle \textbf{T}ransfer (\textbf{HiCAST}), which is able to explicitly exploit clues from multiple semantic levels to customize the stylization results. First, we present a diffusion-based model for image AST, built based on the prevailing architecture of LDM \cite{rombach2022high}. Specifically, we employ the well-trained Stable Diffusion (SD) as backbone model. To inject style information into content image, we integrate content and style images into LDM model and treat them as conditions. For model training, we simultaneously leverage content loss and style loss to preserve the the content coherence meanwhile introduce the fine-grained style details. Second, to realize more flexible and customized AST, we present a new control module, dubbed \textit{Style Adapter}, which is directly pluged into the above model and used to learn the alignment between the multi-level style information and intrinsic knowledge in LDM. As illustrated in Figure \ref{fig:figure1}, we can train various adapters according to different conditions, and their existence allows us to customize the stylization results by freely adjusting combining weights of them. Lastly, we further extend our proposed AST model to process video, via inserting the temporal relation capturing layers into base model. Moreover, we elaborately design an effective learning objective that ensures the output video not only preserve the original stylization effect but also keep satisfactory cross-frame temporal consistency. 

To sum up, our main contributions are summarized as follows: 	
\begin{itemize}
    \item We make first attempt to investigate how to explicitly exploit hierarchical style information to precisely customize the stylization results of Diffusion model for AST task. 
    
    \item We introduce an efficient but effective Style Adapter module for more flexible and customized AST. It introduces the multi-level style information into our designed Diffusion-based image AST model, which leverages the learned alignment knowledge to flexibly manipulate the stylization results. 
        
    \item We build the first Diffusion-based architecture for video AST, in which a novel optimization objective dubbed Harmonious Consistency loss is presented. It significantly promotes the cross-frame temporal consistency meanwhile well preserves the stylization effect. 
    
    \item Extensive experiments and comparisons with other
state-of-the-art methods are conducted to verify
the advance of our proposed method on both image and video AST task.
\end{itemize}

%% file: 2-related-work.tex
\section{Related Work}

\subsection{Image Style Transfer}
The early AST task is explored on images. Gatys et al. \cite{gatys2016image} first exploited the hierarchical representation ability of Deep Neural Networks (DNN) for neural style transfer by aligning the relationships of DNN features. Since then, a rich line of methods have been proposed, boosting the development of AST from different perspectives. Specifically, Huang et al. \cite{huang2017arbitrary} proposed an Adaptive Instance Normalization (AdaIN) operation that utilizes the mean and variance of style feature to replace those of content feature. Li et al. \cite{li2019learning} exploited singular value decomposition to whiten and then re-color images. Afterwards, the advance of attention mechanism \cite{xu2015show} inspires more works. Park et al. \cite{park2019arbitrary} presented to leverage the spatial correlations within content features to re-arranged style features. Liu et al. \cite{liu2021adaattn} combined the core idea of \cite{huang2017arbitrary} and \cite{park2019arbitrary}, leading to more harmonious global and local stylization results. To capture more long-range dependency, subsequent researches \cite{wu2021styleformer,deng2022stytr2} deployed transformer encoder intead of CNNs, which produced more visually plausible results. 

\subsection{Video Style Transfer}
Recently, more researchers \cite{ruder2016artistic,chen2017coherent,wang2020consistent} have also been interested in performing video AST. As a fundamental work, Ruder et al. \cite{ruder2016artistic} warped frames from previous time to the current time, consequently utilizing the constraint of temporal consistency to regulate the optimization. Then, Chen et al. \cite{chen2017coherent} designed an RNN structure baseline and performed the warping operation in the feature space. These methods mostly rely on the correctness of estimated inter-frame correlation, such as optical flows or RNNs. Therefore, the estimation error will degrade the stylization quality and resulting in the ghosting artifacts. Moreover, either estimating optical flows or using RNNs is computationally expensive, making it impractical to generate high-resolution or long videos. Distinct from previous methods, we leverage the inter-frame associations between original content frames to boost temporal consistency of stylized frames, which gets rid of the aforementioned issues.

\subsection{Diffusion Model}
With the prevalence of DMs \cite{song2020denoising, dhariwal2021diffusion, ho2022classifier, rombach2022high}, there has been widespread application in various generative tasks, such as image super resolution \cite{saharia2022image}, image inpainting \cite{lugmayr2022repaint}, and point cloud completion \cite{kasten2023point}. As for image-to-text synthesis, Latent Diffusion Model (LDM) \cite{rombach2022high} is a representative work that performs the denoising process in an auto-encoder’s latent space, effectively reducing the required computation burden. However, it is not easy to directly apply LDM to resolve AST problem, because original LDM requires pairwise data for training. To break this limitation, Ruta et al. \cite{ruta2023diff} proposed to introduce style image into the self-attention blocks of the LDM Unet modules, which re-colored content image by the style one. In parallel, Chen et al. \cite{chen2023artfusion} presented to take both content and style images as the condition of LDM that enables free control over the style and content strength. In comparison to them, we have two key characteristics. First, our method can explicitly leverage multi-level style information to precisely customize results, rather than just on the coarse level of stylization strength. Additionally, we further extend our model to handle video data, which contributes the first benchmark built based on DMs for video AST task. 

%% file: 3-Method.tex
\begin{figure}[t]
    \centering
    \includegraphics[width=0.95\linewidth]{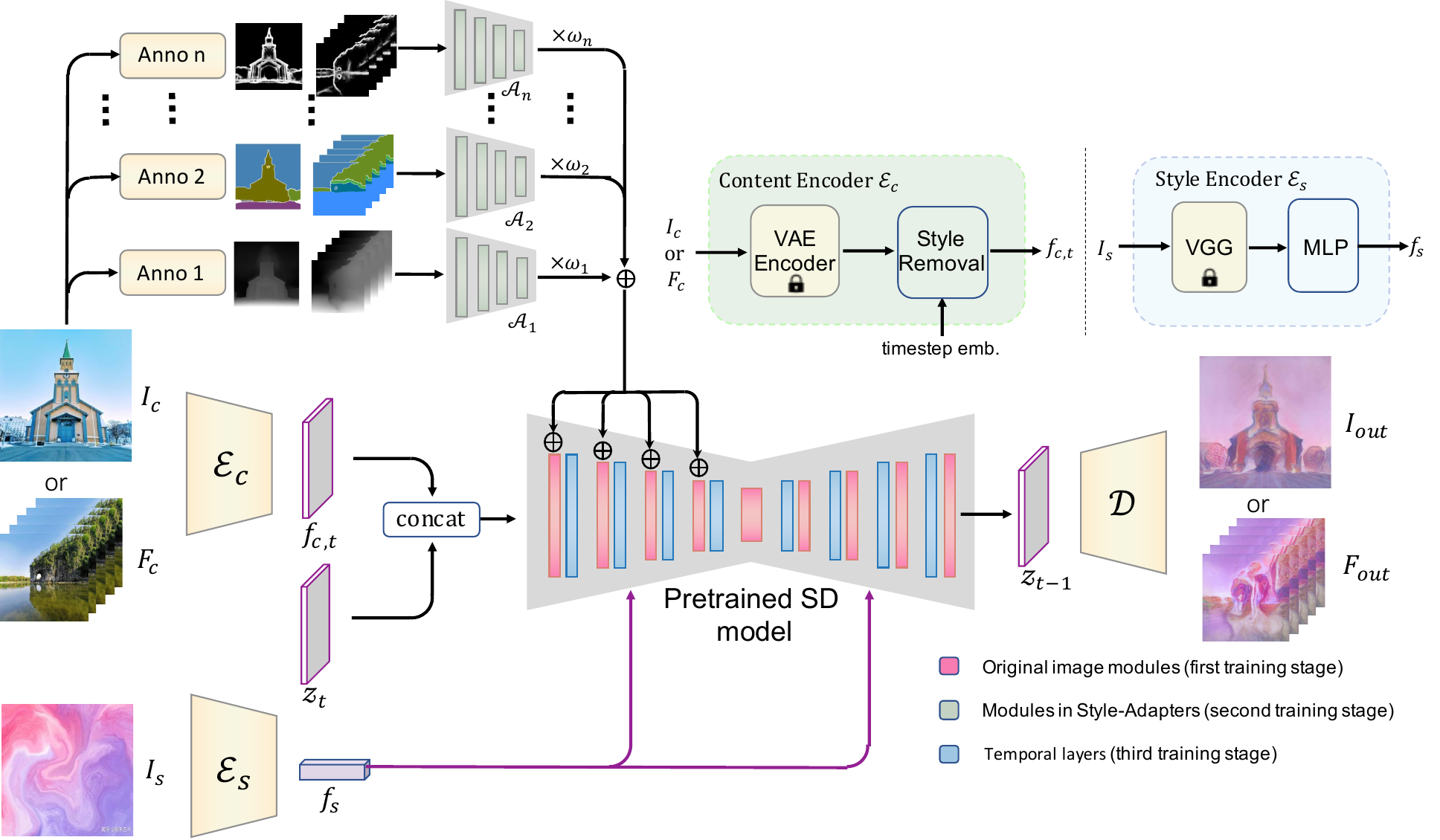}
    \caption{Method overview. Our \algoName firstly utilizes two encoders to extract features from content image $I_c$ or video $F_c$ and style image $I_s$, followed by being fed into our backbone U-Net $\epsilon_\theta$. The U-Net comprises the original image modules and the temporal layers. In the top-left of the figure, the Style-Adapters extract the control maps from the content and then inject the control signals into the U-Net. Finally, the decoder is employed to generate the output $I_{out}$ or $F_{out}$. Our training process is divided into three stages, with the modules represented by the colors red, green, and blue.}
    \vspace{-20pt}
    \label{fig:model_arch}
\end{figure}

\section{Method}

Our proposed model, \algoName, is built on a variant of LDM \cite{rombach2022high}, which is able to explicitly exploit clues from diverse semantic level to customize the stylization results.
In this section, we will introduce the structure and strategies of \algoName, addressing the challenges posed by the lack of unified models for image and video AST. Firstly, we present a meticulously designed network structure tailored for both image and video AST (Section~\ref{sec:3-1}). Subsequently, in Section~\ref{sec:3-2}, we elaborate on a three-stage training strategy developed to optimize the performance of our model. Finally, we outline an inference strategy in Section~\ref{sec:3-3}, allowing for fine-grained control in multiple dimensions during the style transfer process.

\subsection{Model Architecture} \label{sec:3-1}

The architecture of our \algoName encompasses various components, including the style encoder $\mathcal{E}_s$, the content encoder $\mathcal{E}_c$, the decoder $\mathcal{D}$, the backbone U-Net $\epsilon_{\theta}(z_t, f_{c,t}, f_{s}, t)$, and the Style-Adapters, as depicted in Fig. \ref{fig:model_arch}. The model takes as input a content image $I_c$ or video $F_c$ and a style image $I_s$, producing stylized results $I_{out}$ or $F_{out}$.
For the decoder $\mathcal{D}$, we employ a pretrained VAE decoder \cite{kingma2013auto, rombach2022high}, and further details on other components are discussed below.

\textbf{Content Encoder.} The content encoder $\mathcal{E}_c$, illustrated in the upper right of Fig. \ref{fig:model_arch}, processes the content image $I_c$ or video $F_c$ using a pretrained VAE encoder \cite{kingma2013auto, rombach2022high} with a downsampling factor of 8. In the following style removal network, timestep embedding adapts VAE features to each timestep. Subsequent CNNs refine these adapted features, yielding the final time-refined content features $f_{c,t}$.

\textbf{Style Encoder.} The style encoder $\mathcal{E}_s$ comprises a pretrained VGG-16 Network \cite{simonyan2014very} trained on ImageNet \cite{deng2009imagenet} and a style extraction module (upper right of Fig. \ref{fig:model_arch}). We firstly get the VGG features by using the pretrained VGG-16 Network and calculate the mean and variance of the different levels features. And then an MLP converts these statistics to the size of the timestep embedding, producing the final style features $f_s$. This facilitates injecting style information into the backbone U-Net \cite{rombach2022high} by adding adjusted style features to the timestep embedding \cite{chen2023artfusion}.

\textbf{Backbone U-Net.} Our backbone U-Net $\epsilon_{\theta}(z_t, f_{c,t}, f_{s}, t)$, situated in the middle of Fig. \ref{fig:model_arch}, is fine-tuned based on pretrained Stable Diffusion models \cite{rombach2022high} and incorporates new modules. For image AST, only the preserved red parts from the original SD U-Net are utilized, which are called as original image modules. For video AST, blue temporal layers are added. Specifically, each frame’s features undergo processing through 2D convolution or spatial attention layers, followed by a trainable 1D convolutional layer or temporal attention for frame modeling \cite{chen2023control}.The details of the temporal layers are provided in the Appendix \ref{sec:6-4}. This approach ensures more consistent video AST results (see Section~\ref{sec:3-2} for training strategy).

The traditional SD model's backbone U-Net input comprises the noisy latent $z_t$, extracted from the VAE encoder \cite{kingma2013auto, rombach2022high} during the training process and sampled from the Gaussian distribution during inference. Due to the characteristics of DMs \cite{sohl2015deep, ho2020denoising}, our input retains the noisy latent $z_t$. However, simultaneously, we aim to inject content and style information into the network. We achieve this by concatenating the final time-refined content features $f_{c,t}$ with the noisy latent $z_t$ and inputting them concurrently into the U-Net. The first convolutional layer of the network is reinitialized with the initial layer of the pretrained U-Net. And for style features $f_s$ are directly added to the timestep embedding to introduce style features.

\begin{wrapfigure}{r}{0.35\linewidth}
  \centering
  \vspace{-20pt}
  \includegraphics[width=\linewidth]{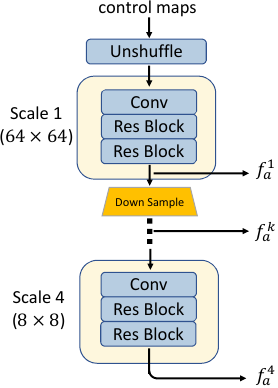}
  \vspace{-20pt}
  \caption{Architecture of the Style-Adapter which includes four scales by downsampling four times.}
  \vspace{-20pt}
  \label{fig:adapter}    
\end{wrapfigure}

\textbf{Style-Adapter.}
To achieve more flexible and personalized AST, we introduce a novel control module called \textit{Style-Adapter}, shown in the upper left of Fig. \ref{fig:model_arch}. This model has the advantage of being lightweight, plug-and-play, composability, and generalizability, which can seamlessly integrates into the backbone U-Net and facilitates the learning of alignment between multi-level style information and intrinsic knowledge in LDM.
Specifically, our Style-Adapter extracts pertinent features from content images or videos and integrates them into the backbone U-Net $\epsilon_{\theta}$. We derive the requisite control maps \cite{zhang2023adding} through distinct annotators, such as edge extraction, semantic segmentation, and depth estimation (if the content input is video, we extract the control maps separately for each frame). Subsequently, these control maps are sent to the adapters, where the features of the control maps are aligned to the same dimension and size as the features of different layers in the backbone U-Net $\epsilon_{\theta}$ (The detailed structure is shown in Fig. \ref{fig:adapter}). Depending on the number of annotators, we can also adjust the number of adapters. We perform a weighted summation of the output features of each adapter and add directly to the backbone U-Net $\epsilon_{\theta}$. By training the backbone network and Style-Adapter separately, we can better adjust the style output we want and achieve flexible and controllable arbitrary style transfer.

\subsection{Training} \label{sec:3-2}
To effectively train the network, the design of appropriate loss functions is crucial. Our training strategy comprises three stages: the image model fine-tune stage, the adapter training stage, and the temporal layers training stage. In each stage, we employ hybrid supervision from both content input and style input, crafting tailored loss functions for training.

\textbf{Image model fine-tune stage.} In this stage, our objective is to train an efficient image style transfer benchmark model based on the pretrained SD model \cite{rombach2022high}. We introduce four supervision methods: content-style supervision, style-style supervision, none-style supervision, and content-none supervision (details provided in the Appendix \ref{sec:6-2}).

Based on the derived hybrid supervision, we define the content loss, style loss, and adversarial loss, constituting the total loss. The content loss $L_c$ serves as a noise-estimate loss, ensuring that the network can reconstruct the content image:
\begin{equation}
    L_c = \lambda_{c} || \epsilon - \epsilon_{\theta}(z_t, f_{c,t}, f_s, t) ||_2
    \label{eq:content}
\end{equation}
where $\epsilon$ is the noise, and $z_t$ denotes the noisy latent at timestep $t$.
For the style loss $L_s$, we firstly obtain the generated result $I_{out}$ by directly putting the estimated noise-free latent $z_0$ into the VAE decoder $\mathcal{D}$ \cite{kingma2013auto, rombach2022high}. And then, VGG features are extracted from both the generated result $I_{out}$ and the original style image $I_s$. Finally, we constrain the mean $\mu$ and standard deviation $\sigma$ of these features \cite{huang2017arbitrary}. The final form of the style loss $L_s$ is given by:
\begin{equation}
L_s = \lambda_{s} (||\mu(\phi(I_{out})) - \mu(\phi(I_{s}))||_2 + ||\sigma(\phi(I_{out})) - \sigma(\phi(I_{s}))||_2)
\end{equation}
where $\phi$ is the pretrained VGG-16 Network \cite{simonyan2014very}. The incorporation of the content loss and style loss ensures that the content information of $I_{out}$ closely resembles $I_c$, while the style information of $I_{out}$ aligns with that of $I_s$. Furthermore, to prevent the model from merely performing simple color transfer, we introduce the adversarial loss \cite{goodfellow2014generative}, which encompasses the base GAN loss $L_G$ and the patch GAN loss $L_{pG}$ \cite{park2020swapping, wang2023pdnlanet}:
\begin{equation}
\begin{aligned}
L_G &= \lambda_{G} (\mathbb{E}[log(D_{\texttt{ori}}(I_{s}))] + \mathbb{E}[1 - log(D_{\texttt{ori}}(I_{out}))]) \\
L_{pG} &= \lambda_{pG} \mathbb{E} [- log( D_{\texttt{patch}}(crop(I_{out}), crops(I_s)))]
\end{aligned}
\end{equation}
where $D_{\texttt{ori}}$ and $D_{\texttt{patch}}$ are the discriminators of $L_G$ and $L_{pG}$.
These two loss functions help us judge whether the entire or each patch of output image is consistent with the style image $I_s$, which can ensure that the details in the stylized output appear perceptually similar, enhancing the overall visual fidelity.

\textbf{Adapter training stage.} In this stage, we maintain the parameters of the baseline image style transfer model unchanged, concentrating solely on the training of our Style-Adapters. With the intention of empowering the Style-Adapters to regulate the output structure, we only utilize the content loss $L_c$ (Eq. \ref{eq:content}) to enhance consistency with the original content in the output. In the inference process, fine-tuned control over the output structure is achieved by judiciously adjusting the weights of the controls. This approach allows for nuanced adjustments, facilitating a more granular control over the style transfer output.

\newcommand{\temploss}{harmonious consistency loss $L_h$\xspace}

\textbf{Temporal layers training stage.}
In this stage, we focus on training the temporal layers in the backbone U-Net $\epsilon_{\theta}$, represented by the blue modules in Fig. \ref{fig:model_arch}. 
To keep better temporal consistency in the output video results, we have designed a dedicated \temploss. 
The \temploss contains two parts: global part and local part.
For the global part, firstly, we assume that the temporal information in the original content video $F_c$ is perfect, while our image baseline model exhibits excellent style transfer capabilities.
Therefore, for the output result by using the network with temporal layers, we constrain it respectively by comparing with the random noise input $\epsilon$ and the noise estimated $\epsilon_{\theta I}$ by the image baseline without temporal layers.
Due to the simplicity of constraining only at the global level, we designed additional local constraints for training.
We utilize a contrastive loss \cite{park2020contrastive} at patch level.
Specifically, for a patch $v$ in the output video, we choose the corresponding patch from the content video as the positive sample $v^+$, and we select the surrounding 8 patches of $v^+$ and 8 non-local patches away from $v^+$ as negative samples $v^-$. 
Our overall \temploss is defined as follows:
\begin{equation}
\begin{gathered}
L_h = \underbrace{ \lambda_{hg1} || \epsilon - \epsilon_{\theta}(z_t, f_{c,t}, f_s, t) ||_2 + \lambda_{hg2} \sum_{i=0}^{F} ||\epsilon_{\theta}^i(z_t,f_{c,t},f_s,t) - \epsilon_{\theta I}(z_t^i,f_{c,t}^i,f_s,t) ||_2}_{\textbf{Global part}} \\
- \underbrace{ \lambda_{hl} \sum \log{\frac{exp(v \cdot v^+ / \tau)}{exp(v \cdot v^+ / \tau) + \sum exp(v \cdot v_n^- / \tau)}}}_{\textbf{Local part}}
\end{gathered}
\end{equation}
where $F$ is the number of the content video $F_c$, $\epsilon_{\theta I}$ is the backbone U-Net without temporal layers, and $\tau$ is the temperature of the contrastive loss.
$z_t$ and $f_{c,t}$ are the noisy latent and time-refined content features of timestep $t$ respectively, and $z_t^i$ and $f_{c,t}^i$ are the i-th frame of them.
By using \temploss, we achieve a substantial enhancement in the coherence and authenticity of our video AST.

\subsection{Inference} \label{sec:3-3}

During the inference process, we initiate the noisy latent $Z_T$ from Gaussian noise, followed by sampling through the DDIM sampler \cite{song2020denoising}. Simultaneously, we acquire the time-refined content features $f_{c,t}$ and the style features $f_s$. When employing Style-Adapters, control maps are extracted from the content image $I_c$ or video $F_c$ through annotators, and then, we extract different levels of features through the Style-Adapters. During each timestep of DDIM sampling, these adapters features are directly added to the backbone U-Net. Fine-grained structural control is achieved by adjusting the weights of different adapter features. After sampling, the estimated noise-free latent $z_0$ is obtained. Utilizing the VAE decoder $\mathcal{D}$ \cite{kingma2013auto, rombach2022high}, we generate the final output image $I_{out}$ or video $F_{out}$.

Additionally, we have incorporated supplementary classifier-free guidance strategies \cite{ho2022classifier}. For the backbone U-Net $\epsilon_{\theta}$, the standard input comprises the time-refined content features $f_{c,t}$, style features $f_s$, and the timestep $t$.  We can leave the two features blank and then perform a weighted sum, defined as:
\begin{equation}
    \hat{\epsilon}_{t} = w_{o} \epsilon_{\theta} (f_{c,t}, f_s, t) + w_{c} \epsilon_{\theta} (f_{c,t}, \varnothing, t) + w_{s} \epsilon_{\theta} (\varnothing, f_s, t)
\end{equation}
where $w_o + w_c + w_s = 1$ (these weights are called style scaling factors). In this way, we achieve fine-grained control of style transfer by adjusting the weights of Style-Adapters and classifier-free guidance.

%% file: 4-experiments.tex
\begin{figure}[t]
    \centering
    \includegraphics[width=\linewidth]{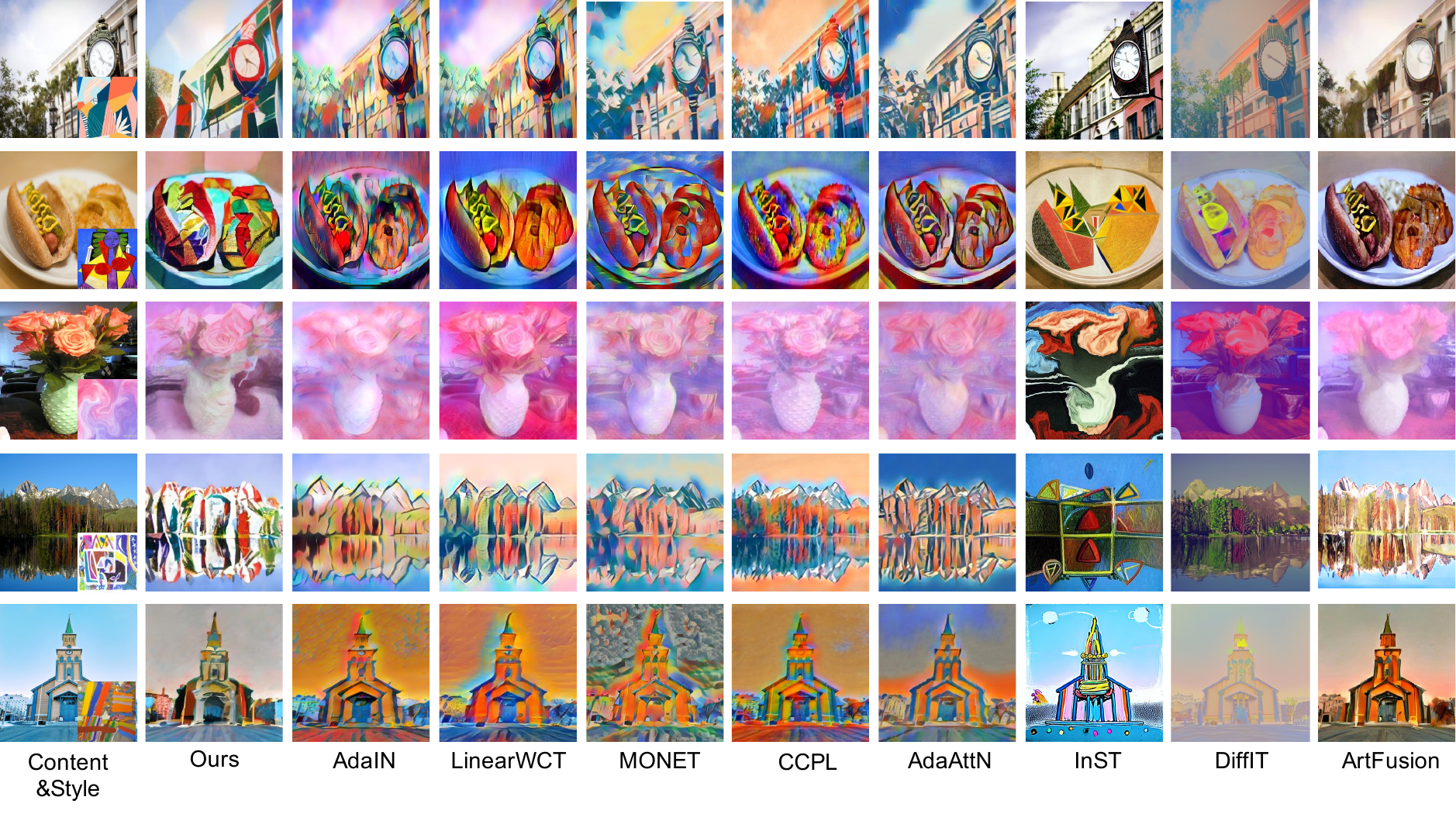}
    \vspace{-25pt}
    \caption{Results of image AST methods.}
    \vspace{-20pt}
    \label{fig:image_results}
\end{figure}

\section{Experiments}

In this part, we first introduce the experimental details and dataset. Afterwards, we list qualitative and quantitative comparisons between our proposed method and several comparison methods. Lastly, we discuss the effect of each component in our model by conducting a series of ablation studies.

\subsection{Dataset and Implementation Details}
\noindent {\textbf{Implementation Details}}
We employ the pretrained Stable Diffusion 1.5 \cite{rombach2022high} as our baseline and fine-tune it accordingly. For the style adapters, we individually train them by using depth maps, semantic segmentation maps, and HED edge maps. More details of our experiment can be found in the Appendix \ref{sec:6-1}.

\noindent {\textbf{Dataset}}
Following existing works \cite{huang2017arbitrary, li2019learning, deng2020arbitrary, chen2021artistic}, on image AST task, we employ MS-COCO \cite{lin2014microsoft} and WikiArt \cite{phillips2011wiki} as the content dataset and style dataset for model training, respectively. For video AST task, we utilize 10k video clips crawled from internet to fine-tune the temporal layers in our model. 

\subsection{Comparison with Image-based Methods}

In this section, for image AST task, we compare our method, HiCAST, with eight representative methods. Five of them are feed-forward AST models: AdaIN \cite{huang2017arbitrary}, LinearWCT(LWCT) \cite{li2019learning}, MANET \cite{deng2020arbitrary}, Adaattn \cite{liu2021adaattn}, CCPL \cite{wu2022ccpl}. The rest are diffusion-based approachs: InST \cite{zhang2023inversion}, DiffIT \cite{kwon2022diffusion}, Artfusion \cite{chen2023artfusion}. 

The subjective results are illustrated in Fig. \ref{fig:image_results}, where the first column depicts our content and style images, the second column showcases the outcomes of our method, the following five columns display the results from feed-forward AST models, and the last three columns exhibit the outcomes of diffusion-based approaches. It is evident that feed-forward AST models tend to emphasize color transfer rather than specific details, resembling color filters rather than style transfer. The previously diffusion-based approaches demonstrate generative capacity, however, their poor controllable ability struggle to transfer the style precisely onto the target rather than the background. 
In contrast, our approach excels in accurately applying style transfer to the target, achieving the most satisfactory subjective results. 

To quantitatively evaluate, we utilize quality score (QS), gram loss \cite{gatys2016image} and LPIPS \cite{zhang2018unreasonable} to evaluate the video style transfer results (Detailed descriptions of the metrics can be found in the Appendix \ref{sec:6-1}). Note that the QS value is averaged by the voting score of 30 participants, in which the score is ranged from 1 to 5 (higher indicates better). It is a subjective metric measuring the stylization effect from perspective of human perception. As shown in Table \ref{tab:image_result}, it can be observed that our approach exhibits a significant improvement over previous methods in terms of the QS metric, while achieving comparable results to other methods in terms of Gram loss and LPIPS.
Our method has achieved a comparable level to feed-forward AST models, surpassing the original diffusion-based methods. In terms of Gram loss and LPIPS, since feed-forward AST models lack generative capacity and tend to directly transfer color, these two metrics are higher. Our method has reached a comparable level to feed-forward AST models, surpassing the previous diffusion-based methods.

\begin{table}[t]
    \centering
    \small
    \vspace{-20pt}
    \caption{Quantitative comparison of image AST methods.}
    \begin{tabular}{lccccc|cccc}
            \toprule
            \multirow{3}{*}{Methods} & \multicolumn{5}{c|}{feed-forward models} & \multicolumn{4}{c}{diffusion-based models} \\
            \cmidrule{2-10}
            & AdaIN  & LWCT  & MANET & AdaAttN & CCPL & InST & DiffIT & Artfusion & \multirow{2}{*}{Ours}\\
            & \cite{huang2017arbitrary} & \cite{li2019learning} & \cite{deng2020arbitrary} & \cite{liu2021adaattn} & \cite{wu2022ccpl} & \cite{zhang2023inversion} & \cite{kwon2022diffusion} & \cite{chen2023artfusion} \\
            \midrule
             QS $\uparrow$ & 2.557 & 2.664 & 2.577 & 2.609 & 2.776 & 2.290 & 2.205 & 2.510 & \textbf{3.369} \\
             Gram loss $\downarrow$ & 0.138 & \textbf{0.103} & 0.191 & 0.152 & 0.191 & 1.110 & 0.704 & 0.302 & 0.153\\
             LPIPS $\downarrow$ & \textbf{0.638} & 0.645 & 0.642 & 0.645 & 0.664 & 0.798 & 0.707 & 0.665 & 0.658\\
            \bottomrule
    \end{tabular}
    \vspace{-10pt}
    \label{tab:image_result}
\end{table}

\subsection{Comparison with Video-based Methods}

\begin{figure}[t]
    \centering
    \includegraphics[width=0.9\linewidth]{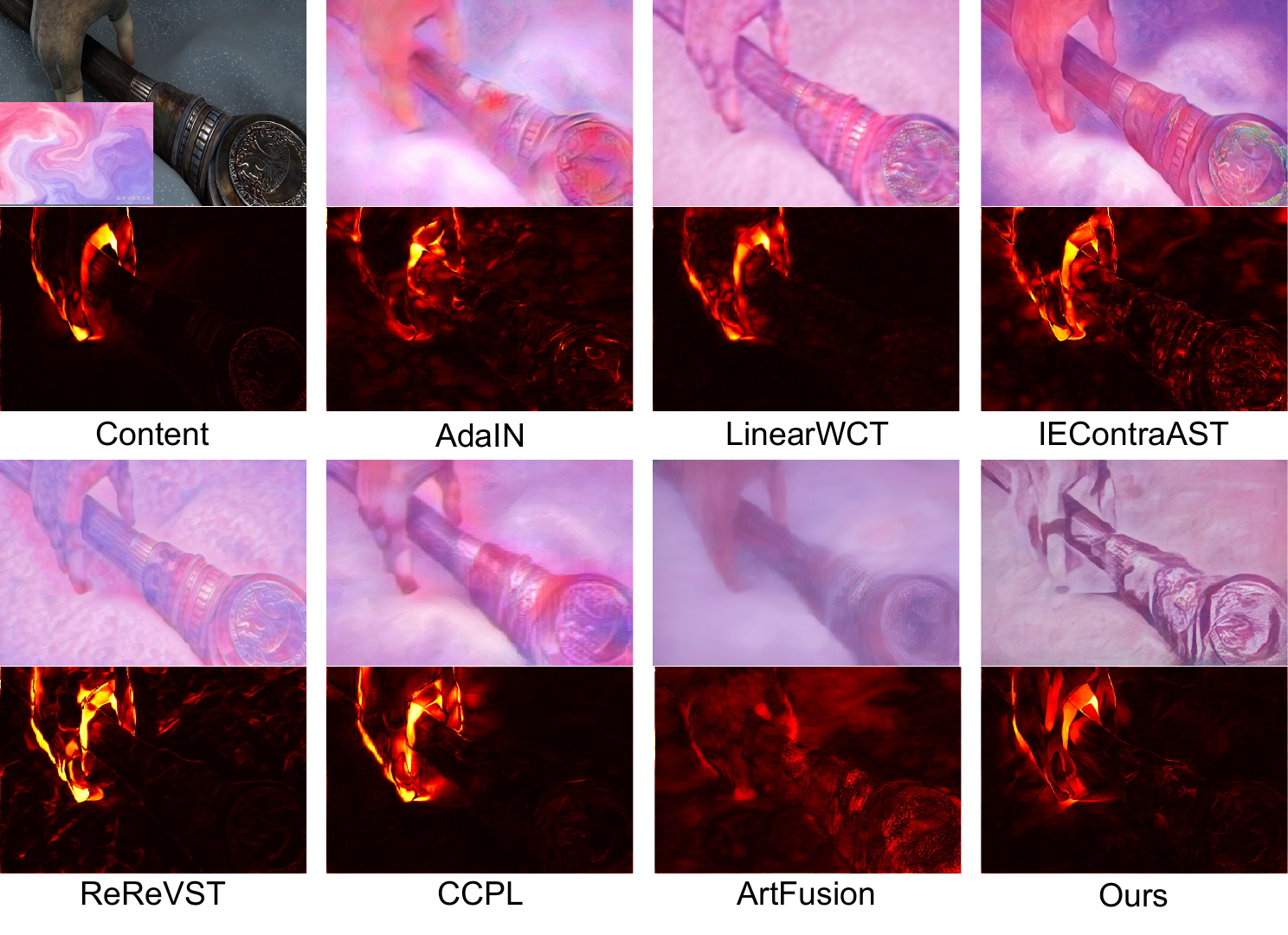}
    \vspace{-10pt}
    \caption{Comparisons of short-term temporal consistency on video AST methods. The odd rows show the previous frame. The even rows show the temporal error heatmap.}
    \label{fig:video_results}
    \vspace{-20pt}
\end{figure}

On video AST task, six leading methods \cite{huang2017arbitrary, li2019learning, chen2021artistic,wang2020consistent, wu2022ccpl, chen2023artfusion} are selected for comparison with our approach. Note that due to \cite{huang2017arbitrary, li2019learning, chen2023artfusion} are single-frame based AST method, their video AST results are conducted between a content video and a style image in a frame-wise manner. 
The subjective results are shown in Fig. \ref{fig:video_results}, where the odd rows show the previous frame, and the even rows show the temporal error heatmap. As we can see, the results of traditional style transfer methods more closely resemble color filters, whereas ours leans more towards generating stylized content. Moreover, with the incorporation of temporal layers and constrained losses, the consistency of our approach can also compete with other methods.

To quantitatively evaluate, we utilize quality score (QS), gram loss \cite{gatys2016image} and temporal loss \cite{wang2020consistent, wu2022ccpl} to evaluate the video style transfer results.
The three evaluation indicators reflect video quality, style transfer degree and video consistency respectively.
The input content videos are from MPI Sintel dataset \cite{butler2012naturalistic} and the style images are from WikiArt dataset \cite{phillips2011wiki}. 
For the temporal loss, we calculate both the short-term consistency ($i=1$) and long-term consistency ($i=10$).
MPI Sintel dataset \cite{butler2012naturalistic} provides ground truth optical flows in the short-term consistency experiments and we use PWC-Net \cite{sun2018pwc} to estimate the optical flow following in the long-term consistency experiments \cite{wang2020consistent, wu2022ccpl}.
Table \ref{tab:video_result} shows that we have achieved better temporal consistency and image quality score, with Gram loss reaching a level comparable to other methods.

\begin{table}[t]
    \centering
    \small
    \vspace{-20pt}
    \caption{Quantitative comparison of video AST methods.}
    \begin{tabular}{lcccc}
            \toprule
            \multirow{2}{*}{Methods} & \multirow{2}{*}{ QS $\uparrow$} & \multirow{2}{*}{Gram loss $\downarrow$} & \multicolumn{2}{c}{Temporal loss $\downarrow$} \\
            \cmidrule{4-5}
            & & & $i=1$ & $i=10$\\
            \midrule
            LWCT \cite{li2019learning} & 2.462 & 0.473 & 0.117 & 0.237 \\
            IEContraAST \cite{chen2021artistic} & 2.821 & 1.062 & 0.141 & 0.262\\
            ReReVST \cite{wang2020consistent} & 2.282 & 0.815 & 0.108 & 0.235\\
            CCPL \cite{wu2022ccpl} & 2.577 & \textbf{0.371} & 0.128 & 0.251\\
            Ours & \textbf{2.846} & 0.482 & \textbf{0.104} & \textbf{0.222} \\
            \bottomrule
    \end{tabular}
    \label{tab:video_result}
    \vspace{-10pt}
\end{table}

\begin{figure}[t]
  \centering
  \includegraphics[width=0.7\linewidth]{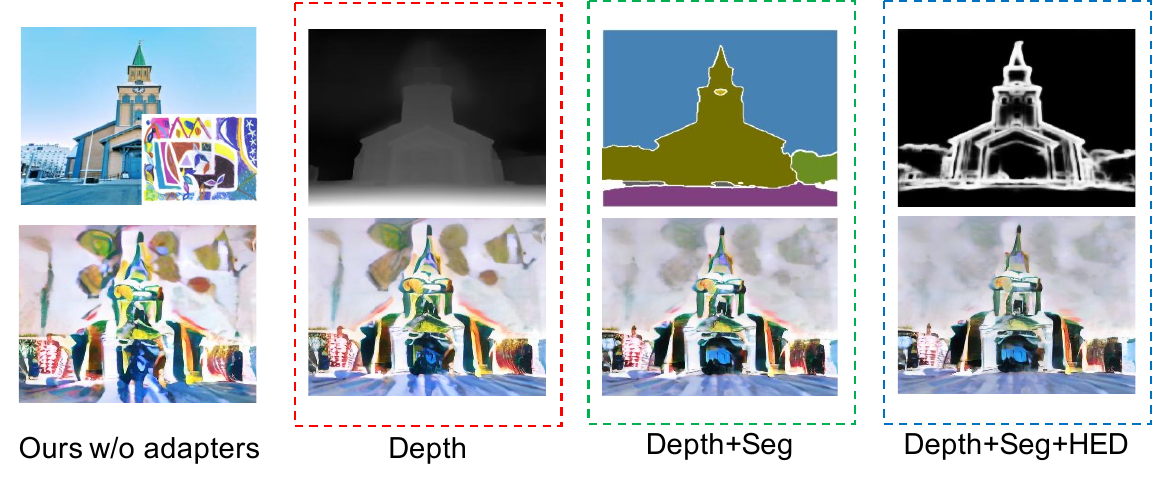}
  \vspace{-10pt}
  \caption{Image AST results with different control maps.}
  \vspace{-10pt}
  \label{fig:abs_control}    
\end{figure}

\begin{figure}[t]
  \centering
  \includegraphics[width=0.85\linewidth]{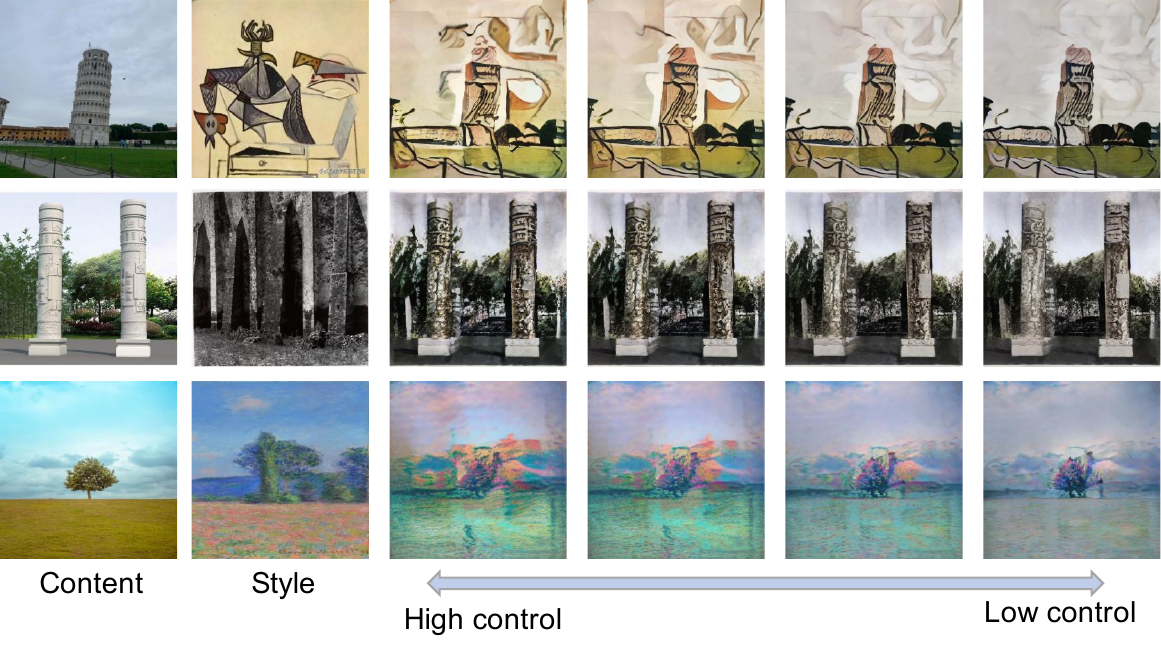}
  \vspace{-10pt}
  \caption{Image AST results using HED controllable maps with different weights.}
  \vspace{-10pt}
  \label{fig:abs_hedweight}    
\end{figure}

\begin{figure}[t]
  \centering
  \includegraphics[width=0.65\linewidth]{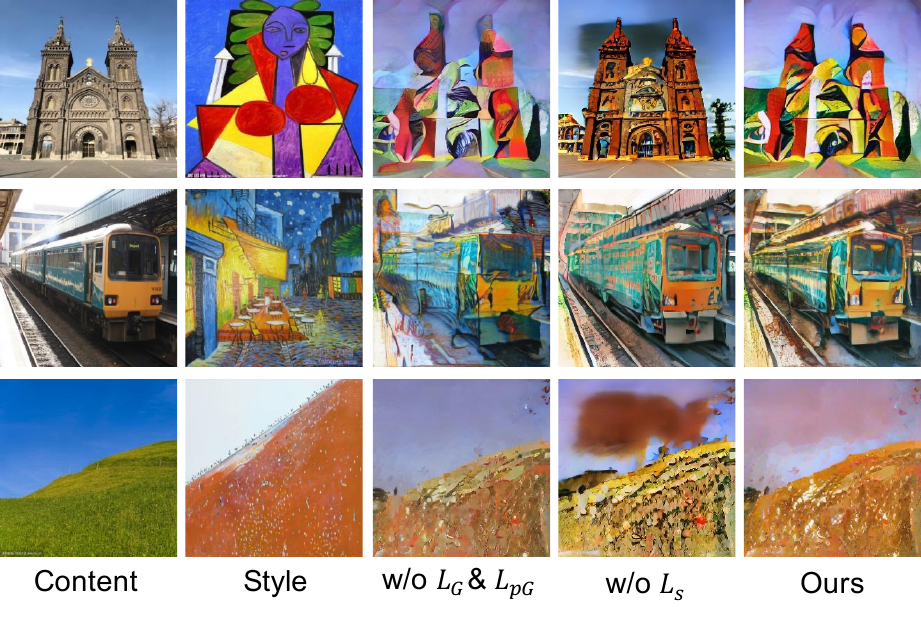}
  \vspace{-10pt}
  \caption{Image AST results with different loss.}
  \vspace{-10pt}
  \label{fig:abs_loss}    
\end{figure}

\begin{figure}[t]
    \centering
    \includegraphics[width=0.9\linewidth]{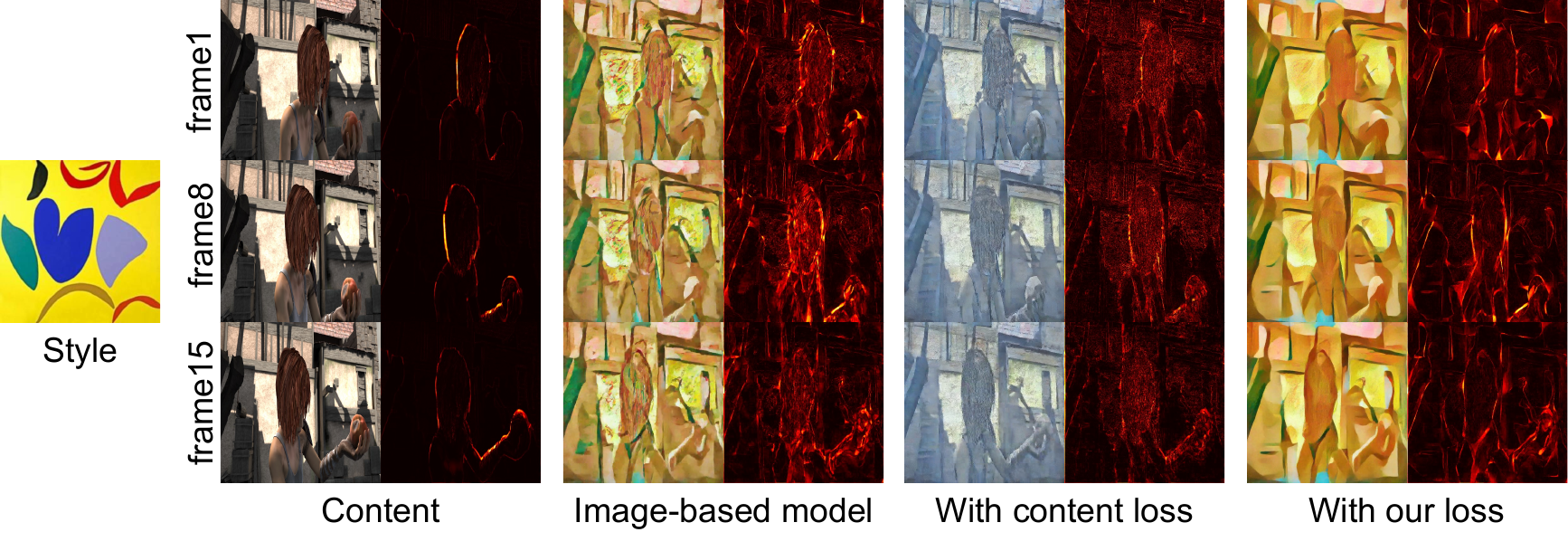}
    \vspace{-10pt}
    \caption{Video AST results with different loss. We show the previous frame and the temporal error heatmap.}
    \label{fig:abs_video}
    \vspace{-10pt}
\end{figure}

\subsection{Ablation Study}
Our ablation study contains two parts: ablation study on image style transfer and on video style transfer. 

\noindent\textbf{Image style transfer} Firstly, we conduct an ablation analysis on adapters, experimenting with different control maps. And then, we perform an ablation analysis on loss functions.
More ablation analysis is shown in the Appendix \ref{sec:6-5}. 

\noindent \textbf{-- On Different Control Maps.}
In addition to controlling the style scaling factors, we leverage different adapters to exert control over the generation outcomes, as depicted in Fig. \ref{fig:abs_control}. We utilize depth maps, semantic segmentation maps, and HED edge maps as control maps, and employing different adapters enables diverse control effects. 
Moreover, adjusting the weights of the output features from adapters allows us to modify the control intensity, as demonstrated in Fig. \ref{fig:abs_hedweight}. Users can customize the generation results by manually tuning the weights according to their preferences.

\noindent \textbf{-- On Different Loss.} And Next, to demonstrate the effectiveness of our style loss and adversarial loss, we conducted an ablation experiment as shown in Fig. \ref{fig:abs_loss}. It is evident that the style loss tends to transfer color, while the adversarial loss leans towards migrating block information. The optimal results are achieved when both losses are employed simultaneously.

\noindent \textbf{Video style transfer} We compared three different designs: generating each frame directly using the image style transfer network, training the network using noise estimation loss as original diffusion models \cite{ho2020denoising}, and using our harmonious consistency loss $L_h$, as shown in Figure \ref{fig:abs_video}. The results generated by the first approach exhibit poor consistency, and solely constraining noise output leads to a deterioration in stylization. Our designed losses effectively enhance inter-frame consistency while preserving stylization, thereby improving the overall results.

%% file: 5-conclusion.tex
\section{Conclusions}

In this paper, we propose a Highly Controllable Arbitary Style Transfer (HiCAST) model for arbitrary style transfer (AST). Specifically, we design an elaborate diffusion-based model, which aligns the external knowledge with the capabilities hided in diffusion models. A novel module namely Style Adapter is presented, which allows us to to explicitly leverage various semantic signal to customize the stylization results. Moreover, a new harmonious consistency loss is devised. Based on it, we construct the first diffusion-based video AST model and make it practicable to simultaneously maintain the stylization effect and obtain satisfactory inter-frame temporal consistency. For evaluation, our extensive experimental results show that our proposed framework has superior performance compared to the existing methods. In particular, we show how the different combination of style adapters affect the output results, which confirms our original conjecture.

%% file: 6-appendix.tex
\section{Appendix}

\subsection{More Implementation Details} \label{sec:6-1}
We employ the pretrained Stable Diffusion 1.5 \cite{rombach2022high} as our baseline and fine-tune it accordingly. For the style adapters, we individually train them by using depth maps \cite{ranftl2020towards}, semantic segmentation maps \cite{caesar2018coco}, and HED edge maps \cite{xie2015holistically}. How to obtain these control maps is depicted in ControlNet \cite{zhang2023adding}.
Concerning the content encoder, we employ a pretrained VAE encoder \cite{kingma2013auto, rombach2022high} to transform a $3\times512\times512$ image into a $4\times64\times64$ latent code, ultimately yielding $3\times64\times64$ content features. As for the style encoder, we use a pretrained VGG-16 Network \cite{simonyan2014very} to compute mean and variance, generating a 2944-dimensional feature vector, which is then transformed to match the timestep embedding size of 1280. Our sampler utilizes the DDIM sampler \cite{song2020denoising} with 20 sampling timesteps. In the image model fine-tune stage, the loss weights for the content loss $L_c$, the style loss $L_s$, the base GAN loss $L_G$, and the patch GAN loss $L_{pG}$ \cite{park2020swapping, wang2023pdnlanet} are set to $\lambda_{c} = 2$, $\lambda_{s} = 4$, $\lambda_{G} = 1$, and $\lambda_{pG} = 2$, respectively. In the temporal layers training stage, the weights for the global part and local part of the harmonious consistency loss $L_h$ are set to $\lambda_{hg1} = 0.01$, $\lambda_{hg2} = 2$ and $\lambda_{hl} = 1$, respectively.
Finally, our evaluation metrics primarily include quality score (QS), LPIPS\cite{zhang2018unreasonable}, Gram loss \cite{gatys2016image}, and temporal loss\cite{wang2020consistent, wu2022ccpl}:

\noindent \textbf{--Quality Score (QS)} For both image and video style transfer results, we created a survey for manual scoring by human evaluators. This metric reflects human subjective evaluation result, which is averaged by the voting score of 30 participants. The score is ranged from 1 to 5 (higher indicates better).

\noindent \textbf{--LPIPS}\cite{zhang2018unreasonable} Perceptual loss is important in vision tasks \cite{johnson2016perceptual, lucas2019generative, liang2023image}, which can reflect the similarity of images. 
In image AST task, we calculate the perceptual similarity between the style image and the output image.

\noindent \textbf{--Gram Loss}\cite{gatys2016image} We extract features from both the style image and the output results using the pretrained VGG-16 Network \cite{simonyan2014very} and calculate the Gram matrix. Subsequently, we compute the Mean Squared Error (MSE) loss for both. This metric reflects the level of stylization in the images.

\noindent \textbf{--Temporal Loss}\cite{wang2020consistent, wu2022ccpl} In video style transfer, this metric reflects the temporal consistency of the video.

\subsection{Temporal layers} \label{sec:6-4}

\begin{wrapfigure}{r}{0.5\linewidth}
  \centering
  \vspace{-20pt}
  \includegraphics[width=\linewidth]{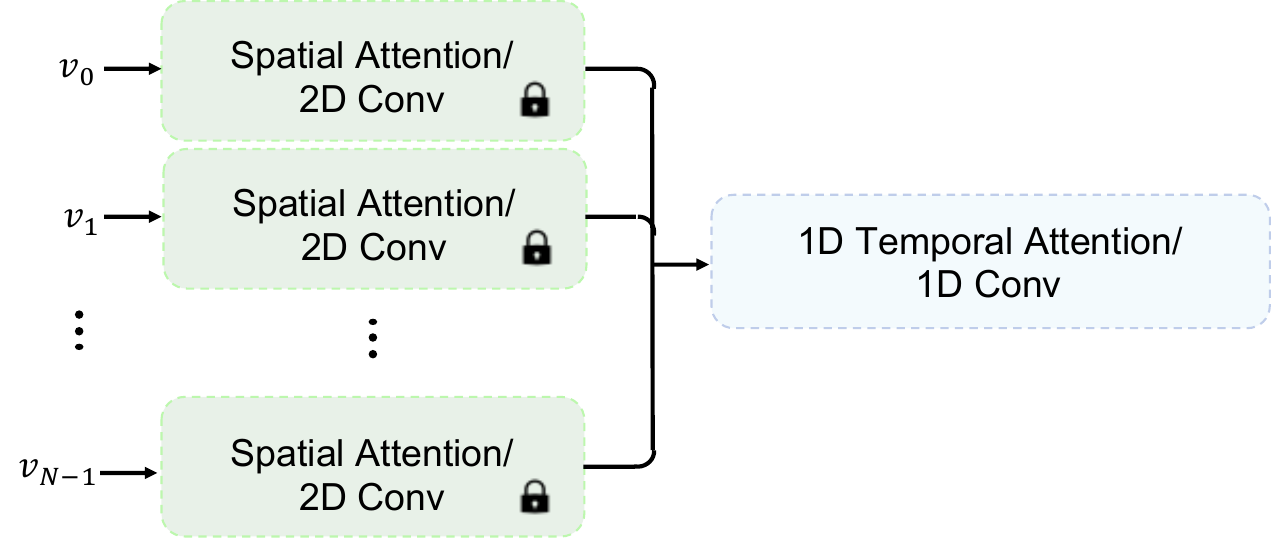}
  \vspace{-20pt}
  \caption{The details of temporal layers. The 2D convolution and the spatial attention layers are frozen. The 1D convolutional layer and temporal attention are training.}
  \vspace{-20pt}
  \label{fig:temporal}    
\end{wrapfigure}

\begin{figure}[t]
  \centering
  \includegraphics[width=0.8\linewidth]{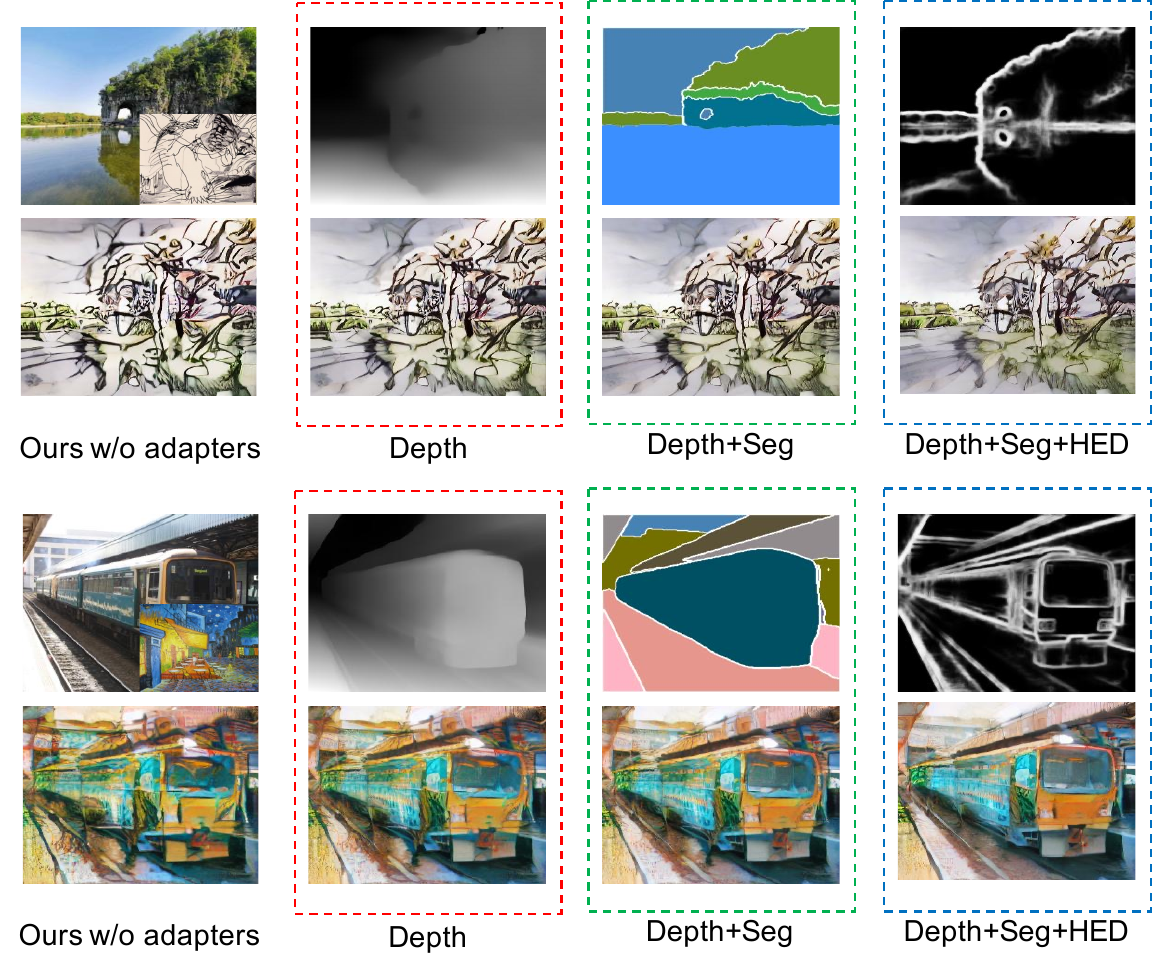}
  \caption{Image AST results with different control maps.}
  \label{fig:sub3}    
\end{figure}

For video AST, blue temporal layers are added into the backbone U-Net, which are shown in the middle of Fig. \ref{fig:model_arch}.
Specifically, each frame’s features undergo processing through 2D convolution or spatial attention layers, followed by a trainable 1D convolutional layer or temporal attention for frame modeling \cite{chen2023control}.
The details of the temporal layers are depicted in Fig. \ref{fig:temporal}. 
Additionally, to facilitate fine-grained modeling, we enhance the spatial self-attention mechanism by integrating spatial-temporal self-attention across frames, formulated as follows:
\begin{equation}
\begin{gathered}
    Attn(Q, K, V) = Softmax (\frac{QK^T}{\sqrt{d}})V, \\
    Q = W^Q\bar{v}_i,K=W^K[\bar{v}_0,...,\bar{v}_{N-1}],V=W^V[\bar{v}_0,...,\bar{v}_{N-1}]
\end{gathered}
\end{equation}
where $\bar{v_i}$ is the features of frame $i$.

\subsection{Hybrid Supervision} \label{sec:6-2}
The input of our \algoName includes the content image and the style image, and we flexibly adjust them to control the training process. We introduce four supervision methods: content-style supervision, style-
style supervision, none-style supervision, and content-none supervision. Next we will introduce in detail.

\noindent \textbf{--Content-style Supervision} This is the most normal input where all loss functions operate normally.

\noindent \textbf{--Style-style Supervision} Here, we use the style image as the input for content, similar to the approach in CycleGAN \cite{zhu2017unpaired, wang2023pdnlanet}. When both the content and style inputs are the same style image, our output should resemble the style image itself.

\noindent \textbf{--None-style Supervision} In this case, we set the content image input to be empty, and the output should be any image with the desired style.

\noindent \textbf{--Content-None Supervision} Here, we set the style image input to be empty, and the output should be the content image itself. The last two supervisions are for subsequent classifier-free guidance \cite{ho2022classifier}.

\begin{figure}[t]
  \centering
  \includegraphics[width=0.9\linewidth]{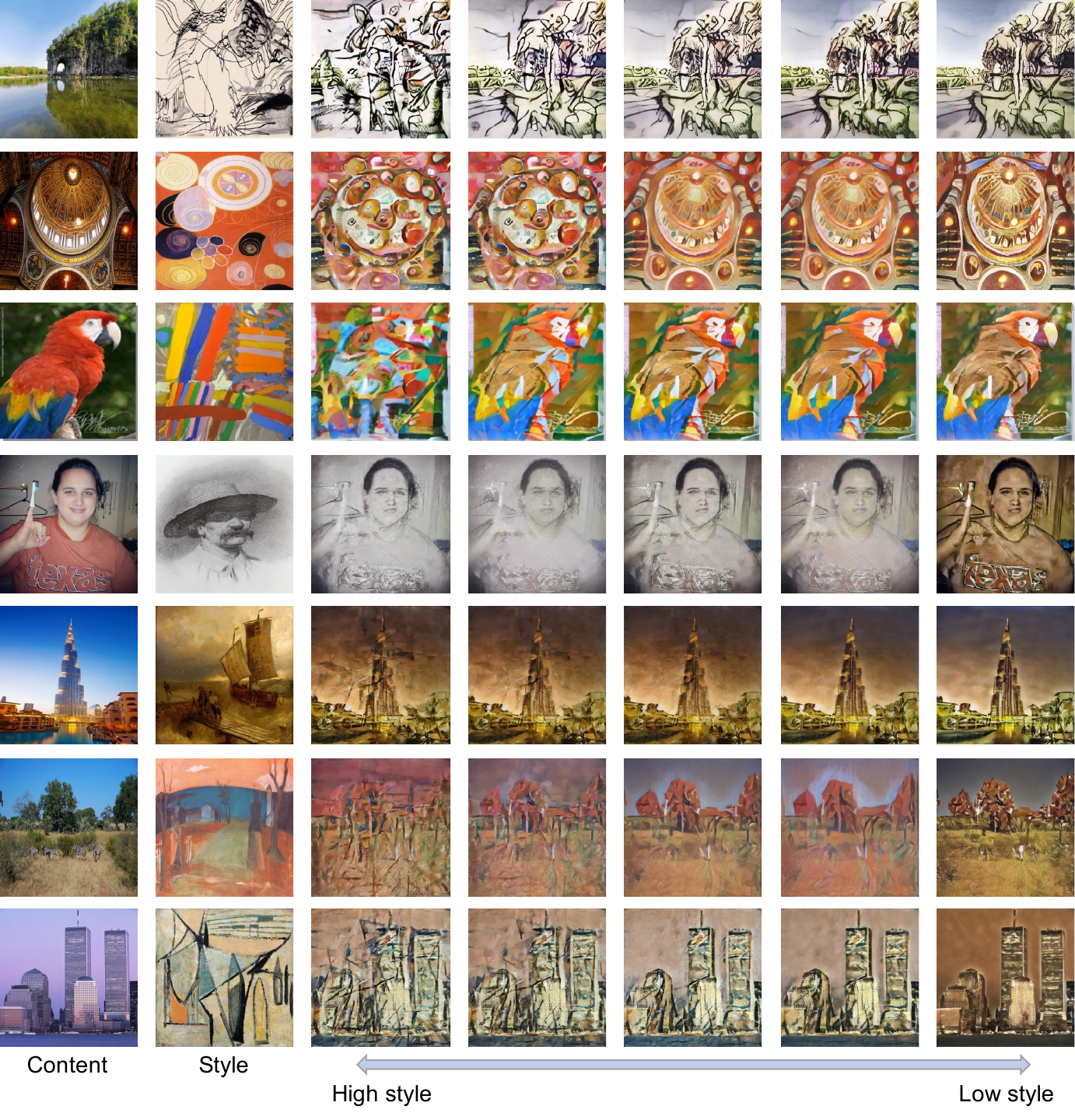}
  \vspace{-10pt}
  \caption{Image AST results with different style scaling factors.}
  \label{fig:abs_style}    
  \vspace{-10pt}
\end{figure}

\subsection{Background} \label{sec:6-3}

The diffusion model (DM) \cite{sohl2015deep, ho2020denoising} has shown impressive performance in various domains as a class of deep generative models, such as image super resolution \cite{saharia2022image}, image inpainting \cite{lugmayr2022repaint}, 3D creation \cite{tang2023make}, and point cloud completion \cite{kasten2023point}.
This approach involves the incorporation of two Markov chains: a forward diffusion chain and a reverse denoising chain. By introducing random perturbations at each time step, the noise gradually diminishes, resulting in the generation of high-quality samples.

Our proposed model, \algoName, is based on a variant of latent diffusion models (LDM) \cite{rombach2022high}. LDM operates within a two-stage framework that incorporates a Variational autoEncoder (VAE) \cite{kingma2013auto} and a diffusion backbone \cite{sohl2015deep, ho2020denoising}. 
For the VAE, it is used to reduce the spatial dimensionality of the image while retaining its semantic essence.
We can convert high-dimensional images into a compact, low-dimensional latent space.
The diffusion backbone operates within this latent space, eliminating the need to manipulate redundant data in the high-dimensional pixel space and thus alleviating the computational burden.

Let $\mathcal{E}$ and $\mathcal{D}$ denote the encoder and decoder of the first-stage VAE, respectively. Consider an image $I$, and let $\epsilon_{\theta}$ represent the diffusion backbone. LDM can be conceptualized as a series of denoising autoencoders $\epsilon_{\theta}(z_t, t)$ for $t=1,...,T$. The training objective aims to predict the noise at timestep $t$ and generate a less noisy version, $z_{t-1}$. Here, $z_{t}$ is obtained from a diffusion process on $z_0 = \mathcal{E}(I)$, modeled as a Markov Chain of length $T$, where each step involves a slight Gaussian perturbation of the preceding state.
By employing the reweighted variational lower bound \cite{dhariwal2021diffusion}, the objective of LDM becomes:

\begin{equation}
    L_{LDM} = \mathbb{E}_{z,\epsilon \sim N(0,I),t \sim U(1,...,T)} [|| \epsilon - \epsilon_{\theta}(z_t,t) ||_2^2]
\end{equation}

Additionally, to achieve conditional image generation, the LDM incorporates a cross-attention mechanism into the U-Net backbone network.
Our \algoName is improved based on the LDM, which injects the content and style information into the model.

\subsection{More Ablation Study} \label{sec:6-5}

\begin{figure}[t]
  \centering
  \includegraphics[width=\linewidth]{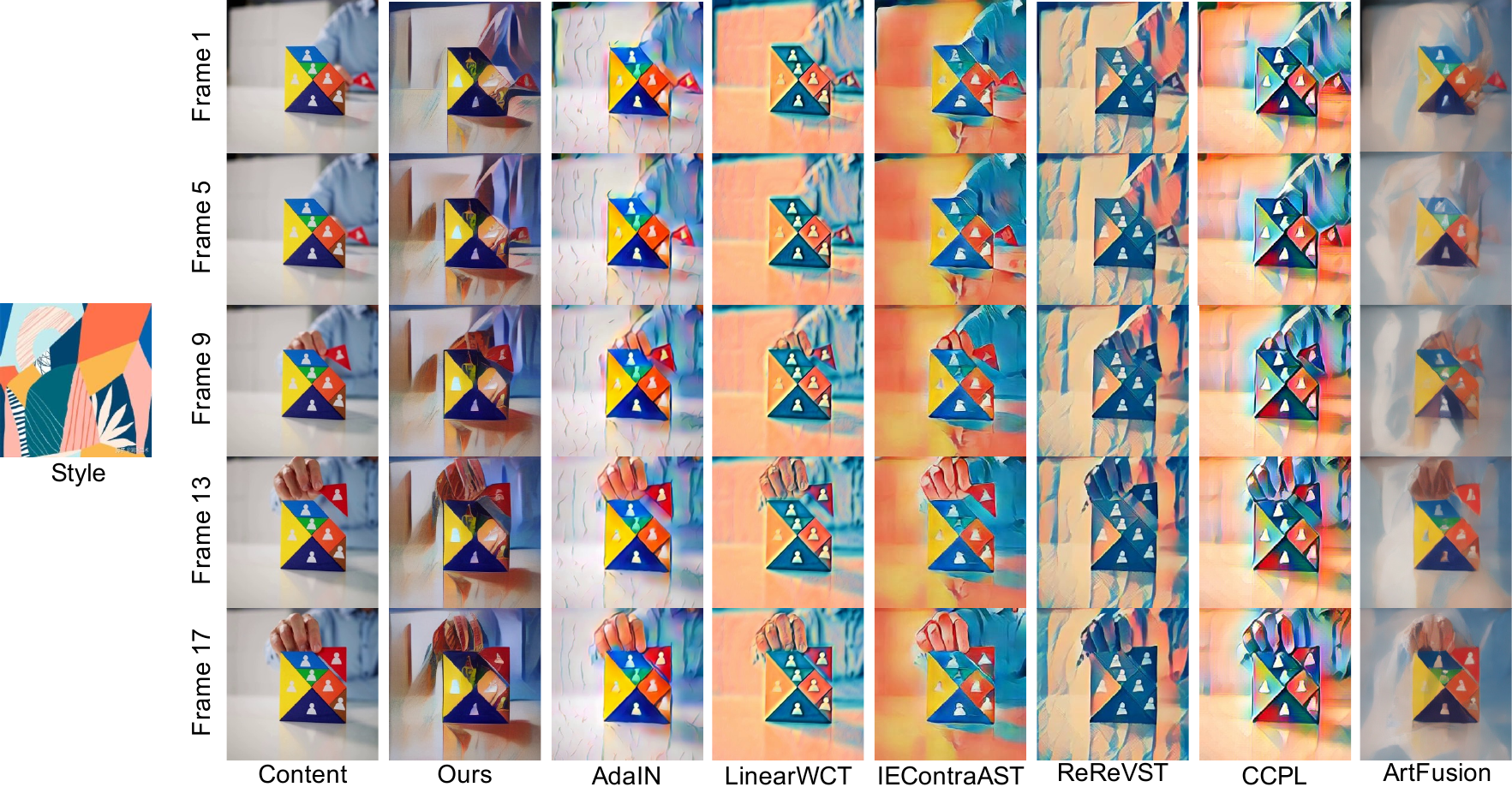}
  \caption{More video AST results of several frames.}
  \label{fig:sub2}    
\end{figure}

Our ablation study contains two parts: ablation study on image style transfer and on video style transfer. 
For image style transfer, we make additional ablation analysis on different style scaling factors.
we verify our ability to achieve varying degrees of style transfer results by adjusting weights under different style scaling factors.
To control the degree of stylization, we employ a classifier-free guidance approach \cite{ho2022classifier} and adjust different style scaling factors, as illustrated in Fig. \ref{fig:abs_style}. The left column represents results with a stronger style, while the right column showcases results with a milder style. Users can manually adjust the weights to select the desired generation outcome.
More results on different control maps are shown in Fig. \ref{fig:sub3}.

\subsection{More Results}

\begin{figure}[t]
  \centering
  \includegraphics[width=\linewidth]{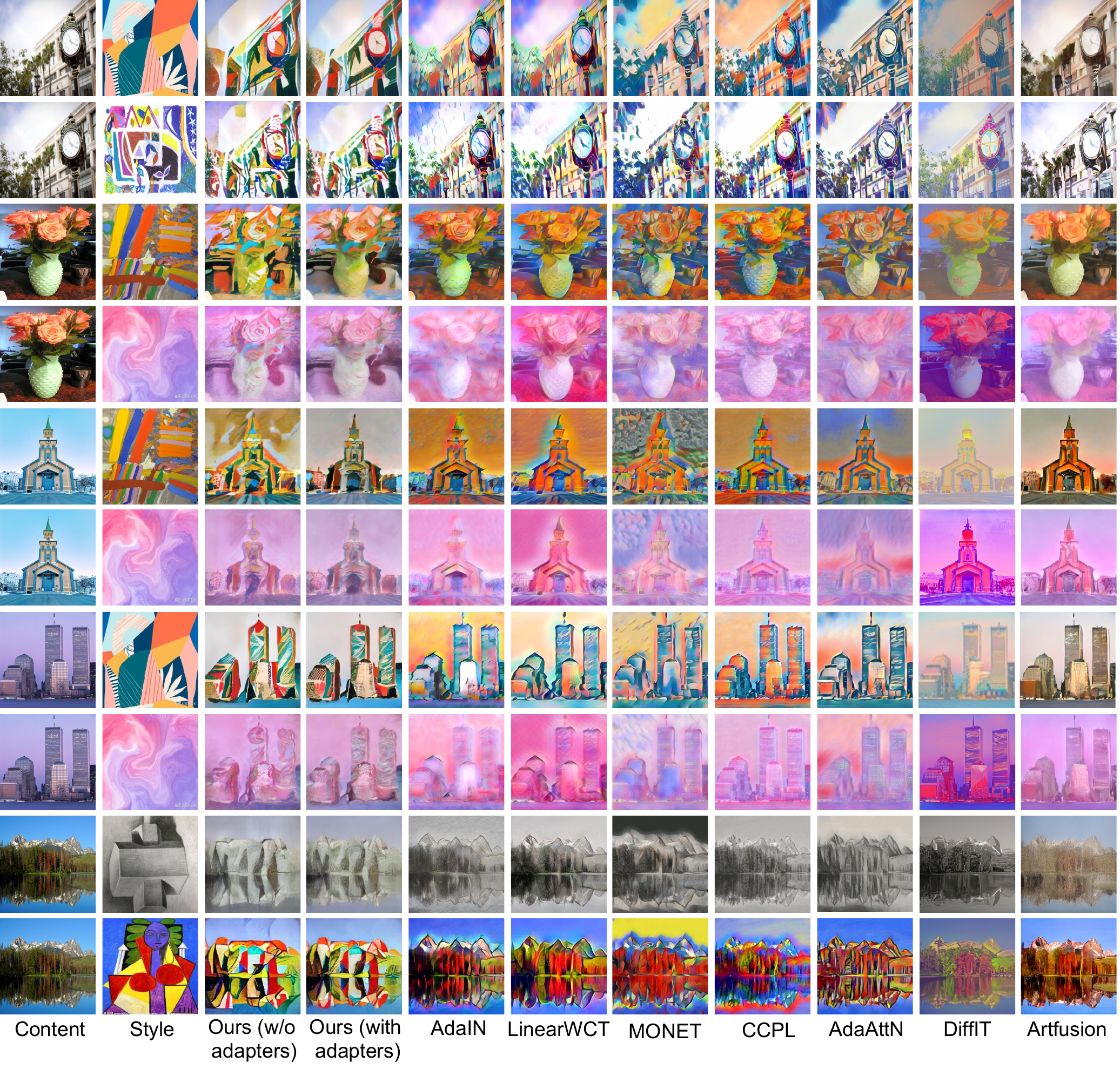}
  \caption{More image AST results. We choose two different style images for each content image.}
  \label{fig:sub1}    
\end{figure}

In this section, we present more results on image AST task and video AST task. 
For images, we perform style transfer on each content image using two different style images (shown in Fig. \ref{fig:sub1}). For videos, we showcase keyframes from the stylized video (shown in Fig. \ref{fig:sub2}).

%% file: main.bbl
\begin{thebibliography}{10}
\providecommand{\url}[1]{\texttt{#1}}
\providecommand{\urlprefix}{URL }
\providecommand{\doi}[1]{https://doi.org/#1}

\bibitem{an2021artflow}
An, J., Huang, S., Song, Y., Dou, D., Liu, W., Luo, J.: Artflow: Unbiased image style transfer via reversible neural flows. In: Proceedings of the IEEE/CVF Conference on Computer Vision and Pattern Recognition. pp. 862--871 (2021)

\bibitem{brempong2022denoising}
Brempong, E.A., Kornblith, S., Chen, T., Parmar, N., Minderer, M., Norouzi, M.: Denoising pretraining for semantic segmentation. In: Proceedings of the IEEE/CVF conference on computer vision and pattern recognition. pp. 4175--4186 (2022)

\bibitem{butler2012naturalistic}
Butler, D.J., Wulff, J., Stanley, G.B., Black, M.J.: A naturalistic open source movie for optical flow evaluation. In: Computer Vision--ECCV 2012: 12th European Conference on Computer Vision, Florence, Italy, October 7-13, 2012, Proceedings, Part VI 12. pp. 611--625. Springer (2012)

\bibitem{caesar2018coco}
Caesar, H., Uijlings, J., Ferrari, V.: Coco-stuff: Thing and stuff classes in context. In: Proceedings of the IEEE conference on computer vision and pattern recognition. pp. 1209--1218 (2018)

\bibitem{chen2023artfusion}
Chen, D.Y.: Artfusion: Arbitrary style transfer using dual conditional latent diffusion models. arXiv preprint arXiv:2306.09330  (2023)

\bibitem{chen2017coherent}
Chen, D., Liao, J., Yuan, L., Yu, N., Hua, G.: Coherent online video style transfer. In: Proceedings of the IEEE International Conference on Computer Vision. pp. 1105--1114 (2017)

\bibitem{chen2021artistic}
Chen, H., Wang, Z., Zhang, H., Zuo, Z., Li, A., Xing, W., Lu, D., et~al.: Artistic style transfer with internal-external learning and contrastive learning. Advances in Neural Information Processing Systems  \textbf{34},  26561--26573 (2021)

\bibitem{chen2022analog}
Chen, T., Zhang, R., Hinton, G.: Analog bits: Generating discrete data using diffusion models with self-conditioning. arXiv preprint arXiv:2208.04202  (2022)

\bibitem{chen2023control}
Chen, W., Wu, J., Xie, P., Wu, H., Li, J., Xia, X., Xiao, X., Lin, L.: Control-a-video: Controllable text-to-video generation with diffusion models. arXiv preprint arXiv:2305.13840  (2023)

\bibitem{deng2009imagenet}
Deng, J., Dong, W., Socher, R., Li, L.J., Li, K., Fei-Fei, L.: Imagenet: A large-scale hierarchical image database. In: 2009 IEEE conference on computer vision and pattern recognition. pp. 248--255. Ieee (2009)

\bibitem{deng2022stytr2}
Deng, Y., Tang, F., Dong, W., Ma, C., Pan, X., Wang, L., Xu, C.: Stytr2: Image style transfer with transformers. In: Proceedings of the IEEE/CVF conference on computer vision and pattern recognition. pp. 11326--11336 (2022)

\bibitem{deng2020arbitrary}
Deng, Y., Tang, F., Dong, W., Sun, W., Huang, F., Xu, C.: Arbitrary style transfer via multi-adaptation network. In: Proceedings of the 28th ACM international conference on multimedia. pp. 2719--2727 (2020)

\bibitem{dhariwal2021diffusion}
Dhariwal, P., Nichol, A.: Diffusion models beat gans on image synthesis. Advances in neural information processing systems  \textbf{34},  8780--8794 (2021)

\bibitem{gatys2016image}
Gatys, L.A., Ecker, A.S., Bethge, M.: Image style transfer using convolutional neural networks. In: Proceedings of the IEEE conference on computer vision and pattern recognition. pp. 2414--2423 (2016)

\bibitem{goodfellow2014generative}
Goodfellow, I.J., Pouget-Abadie, J., Mirza, M., Xu, B., Warde-Farley, D., Ozair, S., Courville, A., Bengio, Y.: Generative adversarial networks. Advances in neural information processing systems, pp. 2672–2680  (2014)

\bibitem{ho2020denoising}
Ho, J., Jain, A., Abbeel, P.: Denoising diffusion probabilistic models. Advances in Neural Information Processing Systems  \textbf{33},  6840--6851 (2020)

\bibitem{ho2022classifier}
Ho, J., Salimans, T.: Classifier-free diffusion guidance. arXiv preprint arXiv:2207.12598  (2022)

\bibitem{huang2017arbitrary}
Huang, X., Belongie, S.: Arbitrary style transfer in real-time with adaptive instance normalization. In: Proceedings of the IEEE international conference on computer vision. pp. 1501--1510 (2017)

\bibitem{jing2019neural}
Jing, Y., Yang, Y., Feng, Z., Ye, J., Yu, Y., Song, M.: Neural style transfer: A review. IEEE transactions on visualization and computer graphics  \textbf{26}(11),  3365--3385 (2019)

\bibitem{johnson2016perceptual}
Johnson, J., Alahi, A., Fei-Fei, L.: Perceptual losses for real-time style transfer and super-resolution. In: Computer Vision--ECCV 2016: 14th European Conference, Amsterdam, The Netherlands, October 11-14, 2016, Proceedings, Part II 14. pp. 694--711. Springer (2016)

\bibitem{kasten2023point}
Kasten, Y., Rahamim, O., Chechik, G.: Point-cloud completion with pretrained text-to-image diffusion models. arXiv preprint arXiv:2306.10533  (2023)

\bibitem{kingma2013auto}
Kingma, D.P., Welling, M.: Auto-encoding variational bayes. arXiv preprint arXiv:1312.6114  (2013)

\bibitem{kwon2022diffusion}
Kwon, G., Ye, J.C.: Diffusion-based image translation using disentangled style and content representation. arXiv preprint arXiv:2209.15264  (2022)

\bibitem{li2019learning}
Li, X., Liu, S., Kautz, J., Yang, M.H.: Learning linear transformations for fast image and video style transfer. In: Proceedings of the IEEE/CVF Conference on Computer Vision and Pattern Recognition. pp. 3809--3817 (2019)

\bibitem{liang2023image}
Liang, P., Jiang, J., Liu, X., Ma, J.: Image deblurring by exploring in-depth properties of transformer. arXiv preprint arXiv:2303.15198  (2023)

\bibitem{lin2014microsoft}
Lin, T.Y., Maire, M., Belongie, S., Hays, J., Perona, P., Ramanan, D., Doll{\'a}r, P., Zitnick, C.L.: Microsoft coco: Common objects in context. In: Computer Vision--ECCV 2014: 13th European Conference, Zurich, Switzerland, September 6-12, 2014, Proceedings, Part V 13. pp. 740--755. Springer (2014)

\bibitem{liu2021adaattn}
Liu, S., Lin, T., He, D., Li, F., Wang, M., Li, X., Sun, Z., Li, Q., Ding, E.: Adaattn: Revisit attention mechanism in arbitrary neural style transfer. In: Proceedings of the IEEE/CVF international conference on computer vision. pp. 6649--6658 (2021)

\bibitem{lucas2019generative}
Lucas, A., Lopez-Tapia, S., Molina, R., Katsaggelos, A.K.: Generative adversarial networks and perceptual losses for video super-resolution. IEEE Transactions on Image Processing  \textbf{28}(7),  3312--3327 (2019)

\bibitem{lugmayr2022repaint}
Lugmayr, A., Danelljan, M., Romero, A., Yu, F., Timofte, R., Van~Gool, L.: Repaint: Inpainting using denoising diffusion probabilistic models. In: Proceedings of the IEEE/CVF Conference on Computer Vision and Pattern Recognition. pp. 11461--11471 (2022)

\bibitem{mou2023t2i}
Mou, C., Wang, X., Xie, L., Zhang, J., Qi, Z., Shan, Y., Qie, X.: T2i-adapter: Learning adapters to dig out more controllable ability for text-to-image diffusion models. arXiv preprint arXiv:2302.08453  (2023)

\bibitem{park2019arbitrary}
Park, D.Y., Lee, K.H.: Arbitrary style transfer with style-attentional networks. In: proceedings of the IEEE/CVF conference on computer vision and pattern recognition. pp. 5880--5888 (2019)

\bibitem{park2020contrastive}
Park, T., Efros, A.A., Zhang, R., Zhu, J.Y.: Contrastive learning for unpaired image-to-image translation. In: Computer Vision--ECCV 2020: 16th European Conference, Glasgow, UK, August 23--28, 2020, Proceedings, Part IX 16. pp. 319--345. Springer (2020)

\bibitem{park2020swapping}
Park, T., Zhu, J.Y., Wang, O., Lu, J., Shechtman, E., Efros, A., Zhang, R.: Swapping autoencoder for deep image manipulation. Advances in Neural Information Processing Systems  \textbf{33},  7198--7211 (2020)

\bibitem{phillips2011wiki}
Phillips, F., Mackintosh, B.: Wiki art gallery, inc.: A case for critical thinking. Issues in Accounting Education  \textbf{26}(3),  593--608 (2011)

\bibitem{ranftl2020towards}
Ranftl, R., Lasinger, K., Hafner, D., Schindler, K., Koltun, V.: Towards robust monocular depth estimation: Mixing datasets for zero-shot cross-dataset transfer. IEEE transactions on pattern analysis and machine intelligence  \textbf{44}(3),  1623--1637 (2020)

\bibitem{rombach2022high}
Rombach, R., Blattmann, A., Lorenz, D., Esser, P., Ommer, B.: High-resolution image synthesis with latent diffusion models. In: Proceedings of the IEEE/CVF conference on computer vision and pattern recognition. pp. 10684--10695 (2022)

\bibitem{ruder2016artistic}
Ruder, M., Dosovitskiy, A., Brox, T.: Artistic style transfer for videos. In: Pattern Recognition: 38th German Conference, GCPR 2016, Hannover, Germany, September 12-15, 2016, Proceedings 38. pp. 26--36. Springer (2016)

\bibitem{ruta2023diff}
Ruta, D., Tarr{\'e}s, G.C., Gilbert, A., Shechtman, E., Kolkin, N., Collomosse, J.: Diff-nst: Diffusion interleaving for deformable neural style transfer. arXiv preprint arXiv:2307.04157  (2023)

\bibitem{saharia2022image}
Saharia, C., Ho, J., Chan, W., Salimans, T., Fleet, D.J., Norouzi, M.: Image super-resolution via iterative refinement. IEEE Transactions on Pattern Analysis and Machine Intelligence  \textbf{45}(4),  4713--4726 (2022)

\bibitem{schuhmann2022laion}
Schuhmann, C., Beaumont, R., Vencu, R., Gordon, C., Wightman, R., Cherti, M., Coombes, T., Katta, A., Mullis, C., Wortsman, M., et~al.: Laion-5b: An open large-scale dataset for training next generation image-text models. Advances in Neural Information Processing Systems  \textbf{35},  25278--25294 (2022)

\bibitem{simonyan2014very}
Simonyan, K., Zisserman, A.: Very deep convolutional networks for large-scale image recognition. arXiv preprint arXiv:1409.1556  (2014)

\bibitem{sohl2015deep}
Sohl-Dickstein, J., Weiss, E., Maheswaranathan, N., Ganguli, S.: Deep unsupervised learning using nonequilibrium thermodynamics. In: International Conference on Machine Learning. pp. 2256--2265. PMLR (2015)

\bibitem{song2020denoising}
Song, J., Meng, C., Ermon, S.: Denoising diffusion implicit models. arXiv preprint arXiv:2010.02502  (2020)

\bibitem{srinivasan2021wit}
Srinivasan, K., Raman, K., Chen, J., Bendersky, M., Najork, M.: Wit: Wikipedia-based image text dataset for multimodal multilingual machine learning. In: Proceedings of the 44th International ACM SIGIR Conference on Research and Development in Information Retrieval. pp. 2443--2449 (2021)

\bibitem{sun2018pwc}
Sun, D., Yang, X., Liu, M.Y., Kautz, J.: Pwc-net: Cnns for optical flow using pyramid, warping, and cost volume. In: Proceedings of the IEEE conference on computer vision and pattern recognition. pp. 8934--8943 (2018)

\bibitem{tang2023make}
Tang, J., Wang, T., Zhang, B., Zhang, T., Yi, R., Ma, L., Chen, D.: Make-it-3d: High-fidelity 3d creation from a single image with diffusion prior. arXiv preprint arXiv:2303.14184  (2023)

\bibitem{wang2023pdnlanet}
Wang, H., Zhai, D., Liu, X., Jiang, J., Gao, W.: Unsupervised deep exemplar colorization via pyramid dual non-local attention. IEEE Transactions on Image Processing (TIP)  (2023)

\bibitem{wang2020consistent}
Wang, W., Yang, S., Xu, J., Liu, J.: Consistent video style transfer via relaxation and regularization. IEEE Transactions on Image Processing  \textbf{29},  9125--9139 (2020)

\bibitem{wu2021styleformer}
Wu, X., Hu, Z., Sheng, L., Xu, D.: Styleformer: Real-time arbitrary style transfer via parametric style composition. In: Proceedings of the IEEE/CVF International Conference on Computer Vision. pp. 14618--14627 (2021)

\bibitem{wu2022ccpl}
Wu, Z., Zhu, Z., Du, J., Bai, X.: Ccpl: contrastive coherence preserving loss for versatile style transfer. In: European Conference on Computer Vision. pp. 189--206. Springer (2022)

\bibitem{xie2015holistically}
Xie, S., Tu, Z.: Holistically-nested edge detection. In: Proceedings of the IEEE international conference on computer vision. pp. 1395--1403 (2015)

\bibitem{xu2015show}
Xu, K., Ba, J., Kiros, R., Cho, K., Courville, A., Salakhudinov, R., Zemel, R., Bengio, Y.: Show, attend and tell: Neural image caption generation with visual attention. In: International conference on machine learning. pp. 2048--2057. PMLR (2015)

\bibitem{zhang2023adding}
Zhang, L., Rao, A., Agrawala, M.: Adding conditional control to text-to-image diffusion models. In: Proceedings of the IEEE/CVF International Conference on Computer Vision. pp. 3836--3847 (2023)

\bibitem{zhang2018unreasonable}
Zhang, R., Isola, P., Efros, A.A., Shechtman, E., Wang, O.: The unreasonable effectiveness of deep features as a perceptual metric. In: Proceedings of the IEEE conference on computer vision and pattern recognition. pp. 586--595 (2018)

\bibitem{zhang2023inversion}
Zhang, Y., Huang, N., Tang, F., Huang, H., Ma, C., Dong, W., Xu, C.: Inversion-based style transfer with diffusion models. In: Proceedings of the IEEE/CVF Conference on Computer Vision and Pattern Recognition. pp. 10146--10156 (2023)

\bibitem{zhou20213d}
Zhou, L., Du, Y., Wu, J.: 3d shape generation and completion through point-voxel diffusion. In: Proceedings of the IEEE/CVF International Conference on Computer Vision. pp. 5826--5835 (2021)

\bibitem{zhu2017unpaired}
Zhu, J.Y., Park, T., Isola, P., Efros, A.A.: Unpaired image-to-image translation using cycle-consistent adversarial networks. In: Proceedings of the IEEE international conference on computer vision. pp. 2223--2232 (2017)

\end{thebibliography}
